\pdfoutput=1

\documentclass[11pt]{article}

\usepackage[]{ACL2023}

\usepackage{times}
\usepackage{latexsym}

\usepackage[T1]{fontenc}

\usepackage[utf8]{inputenc}

\usepackage{microtype}

\usepackage{inconsolata}

\usepackage{graphicx}
\usepackage{pgf}
\usepackage{float}
\usepackage{listings}
\usepackage{xcolor}
\usepackage{todonotes}
\usepackage{booktabs}
\usepackage{multirow}
\usepackage{amsmath}

\title{Measuring the Groundedness of Legal Question-Answering Systems}

\author{
    Dietrich Trautmann, {\bf Natalia Ostapuk}, {\bf Quentin Grail}, {\bf Adrian Alan Pol}, {\bf Guglielmo Bonifazi}, \\ {\bf Shang Gao} \and {\bf Martin Gajek}\\
    Thomson Reuters Labs, Zug, Switzerland\\
    \texttt{\{first.last\}@tr.com}
}

\begin{document}

\maketitle

\begin{abstract}
In high-stakes domains like legal question-answering, the accuracy and trustworthiness of generative AI systems are of paramount importance.
This work presents a comprehensive benchmark of various methods to assess the groundedness of AI-generated responses, aiming to significantly enhance their reliability.
Our experiments include similarity-based metrics and natural language inference models to evaluate whether responses are well-founded in the given contexts.
We also explore different prompting strategies for large language models to improve the detection of ungrounded responses.
We validated the effectiveness of these methods using a newly created grounding classification corpus, designed specifically for legal queries and corresponding responses from retrieval-augmented prompting, focusing on their alignment with source material.
Our results indicate potential in groundedness classification of generated responses, with the best method achieving a macro-F1 score of $0.8$.
Additionally, we evaluated the methods in terms of their latency to determine their suitability for real-world applications, as this step typically follows the generation process.
This capability is essential for processes that may trigger additional manual verification or automated response regeneration.
In summary, this study demonstrates the potential of various detection methods to improve the trustworthiness of generative AI in legal settings.
\end{abstract}


\section{Introduction}

\begin{figure}[htbp]
  \centering
  \includegraphics[width=\columnwidth]{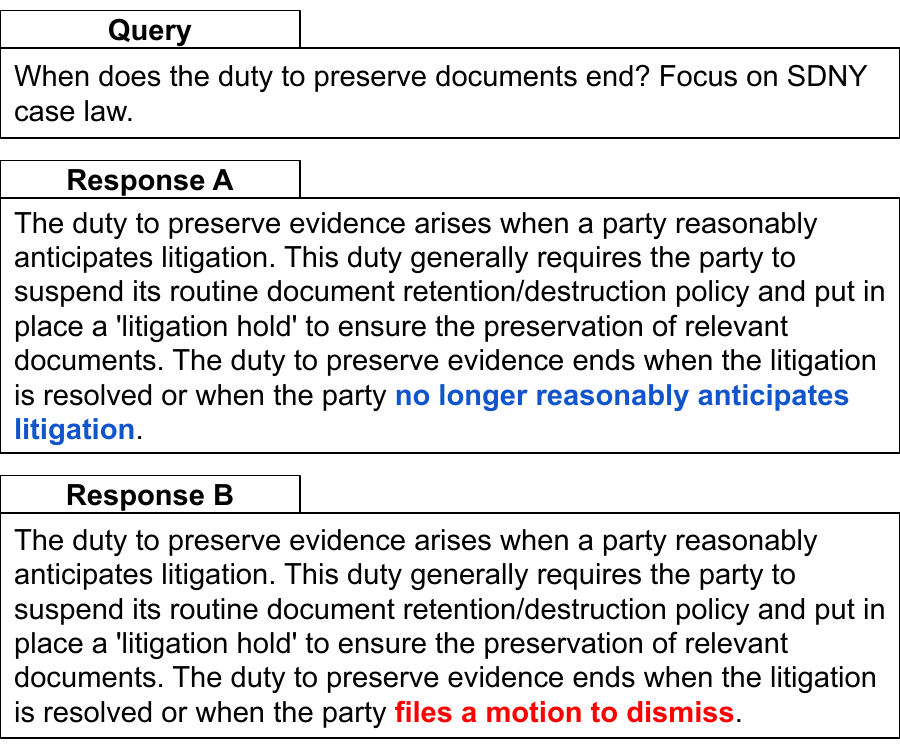}
  \caption{Example query and corresponding LLM responses with \textcolor{blue}{\textbf{grounded}} and \textcolor{red}{\textbf{erroneous}} spans (Procedural Errors). The retrieved context used for grounding the responses was omitted due to its length. The remaining sentences in both responses are identical and grounded, but not highlighted to emphasize the differences.}
  \label{fig:example}
\end{figure}


Generative AI systems are increasingly employed in high-stakes domains such as legal question-answering, where accuracy and trust are paramount \cite{monroy2009nlp,vold2021using,khazaeli2021free,martinez2023survey}.
A significant challenge in these applications is the detection of outputs that are not grounded in the input data (context), which can compromise user trust and diminish the application's value \cite{maynez-etal-2020-faithfulness,rawte2023survey}.
This work addresses this challenge by conducting a comprehensive benchmarking to assess the groundedness of AI-generated legal responses, thereby enhancing their reliability. 

Our methodology investigates diverse approaches to classify responses based on their foundation in the provided source material (cf. Fig. \ref{fig:example}).
We utilize: 
\begin{enumerate}
    \item \emph{Similarity-based techniques}, employing various text similarity metrics to quantify the alignment between the generated text and the input data at the sentence-level.
    \item \emph{Natural language inference} models to determine if the generated response sentences are entailed by or contradict the sentences in the source material.
    \item Diverse \emph{prompting strategies} for large language models (LLMs) to detect ungrounded responses. \cite{bubeck2023sparks}.
\end{enumerate}

We evaluate these approaches on a new corpus of legal queries and responses, annotated for their degree of groundedness. 

Experimental results demonstrate the effectiveness of many methods in the detection of potentially ungrounded answers.
We also discuss the trade-offs between task performance and computational efficiency, highlighting the capabilities of particular approaches to operate with minimal added latency in real-world applications. 

Furthermore, we investigated the types of errors present in the responses, categorizing them into six distinct classes: \emph{Factual Inaccuracies}, \emph{Contextual Misinterpretations}, \emph{Procedural Errors}, \emph{Reasoning Errors}, \emph{Misattributions}, and \emph{Terminological Errors}.
Our analysis reveals that factual inaccuracies are the most prevalent type of errors.
Importantly, we found that the misclassification rates in the overall groundedness assessment task are not uniform across these error categories, providing valuable insights for targeted improvements in AI-generated legal responses.

Our findings underscore the potential of automated groundedness assessment tools to improve the reliability and utility of generative AI in legal settings, ensuring that the generated responses are consistently accurate and trustworthy.
The error analysis further contributes to a nuanced understanding of the challenges in this domain, paving the way for more refined and effective AI systems in legal applications.

\section{Related Work}

\subsection{Grounding of Generated Responses}

Grounding and factual consistency in language model outputs, especially for summarization and question-answering tasks, have been a focal point of recent research. \newcite{kryscinski2020evaluating} introduced a weakly-supervised, model-based approach to verify factual consistency between source documents and generated summaries. This method uniquely combines consistency checks with the extraction of supporting and contradictory spans.

Building on this, \newcite{maynez-etal-2020-faithfulness} performed an extensive human evaluation of neural abstractive summarization systems. Their results showed a significant amount of ungrounded content in model-generated summaries and found that textual entailment measures correlate more strongly with faithfulness than standard metrics. This finding closely relates to our interest in assessing the groundedness of AI-generated legal responses.

The Chain-of-Knowledge (CoK) framework \cite{li2023chain} marks a major advance in reducing hallucinations. By dynamically incorporating grounding information from various sources, CoK enhances factual accuracy in knowledge-intensive tasks. 

In essence, grounding of LLM-generated responses aims to ensure that outputs are factually consistent with input data, thereby enhancing reliability and reducing ungrounded LLM-generated content.

\subsection{Hallucination Detection}

Advancements in hallucination detection have been pivotal in developing more reliable and grounded LLMs, particularly for question-answering (QA) systems.

The \emph{HaluEval-Wild} benchmark \cite{zhu2024halueval} offers a novel approach to evaluating LLM hallucinations in real-world settings. By categorizing challenging user queries into five distinct types, this tool provides essential insights for enhancing LLM reliability in scenarios that mirror real-world interactions, which is crucial for QA systems.

\newcite{wang2024llms} contribute with \emph{MIGRES}, a method that uses LLMs' ability to identify missing information for targeted knowledge retrieval and extraction. This approach promises to improve the groundedness of responses by ensuring comprehensive information gathering.

In long-form question answering, \newcite{rosenthal2024clapnq} introduced \emph{ClapNQ}, a benchmark designed for retrieval-augmented generation (RAG) systems. Its emphasis on concise, cohesive answers grounded in source passages makes it particularly relevant for evaluating QA systems that require detailed, well-supported responses.

An empirical evaluation of AI-driven legal research tools \cite{magesh2024hallucination} challenges claims of "hallucination-free" systems, underscoring the necessity for rigorous evaluation in assessing the groundedness of legal QA systems.

Additionally, \newcite{hong2024hallucinations} have launched the \emph{Hallucinations Leaderboard}, an open initiative for measuring and comparing hallucinations across various LLMs and tasks. This resource offers a valuable opportunity for benchmarking the groundedness of QA systems against a diverse range of models and applications.

\section{Grounding Definition}

Grounding in legal question-answering systems refers to the extent to which an AI-generated response is firmly rooted in, supported by, and directly attributable to the provided legal source material. It ensures the model's output aligns with and accurately represents the information in the input data, avoiding fabrication, extraneous details, or misleading content. A well-grounded response should adhere closely to the facts, legal principles, and reasoning presented in the source material, without introducing unsupported claims or misrepresenting the legal context \cite{chandu-etal-2021-grounding}.

Several key aspects ensure the reliability of AI-generated legal responses. Factual alignment and relevance are crucial, ensuring the content reflects the source documents and addresses the legal query accurately. Source attribution allows tracing information back to specific input texts, while legal interpretation fidelity ensures conclusions are substantiated by the provided materials. This involves not only accurately conveying factual information but also maintaining the integrity of legal procedures, correctly interpreting the context, and using appropriate legal terminology. The generated responses must adhere to the given context, avoiding unsupported claims or extrapolations, and preserving the nuances and complexities of legal language and concepts \cite{magesh2024hallucination}.

The assessment of grounding in legal AI responses involves a comprehensive evaluation of how faithfully the generated content aligns with the retrieved legal context. This evaluation considers various aspects of the response, including its factual accuracy, the appropriateness of legal interpretations, the coherence of legal reasoning, and the proper use of legal terminology. Grounding is vital in legal applications to maintain the integrity of legal advice, ensure compliance with laws and precedents, and prevent misinformation. By ensuring strong grounding, legal question-answering systems can provide more reliable, trustworthy, and legally sound responses, which is crucial in the high-stakes environment of legal practice and decision-making.


\begin{table*}
\centering
\begin{tabular}{lrrr}
\hline
\textbf{Split} & \textbf{\#Queries} & \textbf{\#Responses} & \textbf{\#Response Sentences}\\
\hline
Training    & 400 & 1080 & 5671 \\
Development &  58 &  162 &  797 \\
Testing     & 115 &  316 & 1516 \\
\hline
Total       & 573 & 1558 & 7984 \\
\hline
\end{tabular}
\caption{Data Set Statistics}
\label{data_set_statistics}
\end{table*}

\section{Dataset Creation}

In this section, we will describe and list all the steps involved in creating the \emph{Groundedness Classification} dataset used in our benchmarking.

\subsection{Data Source}

The dataset originates from proprietary data in the \emph{Casetext} Legal Research Skill\footnote{\url{https://casetext.com/cocounsel/}}.
We limited the data selection to the internal users only, primarily consisting of diverse sales demonstrations as well as domain experts and engineering-related testing sessions.
All queries, however, are realistic representations of everyday research in the legal domain.
Additionally, we performed a deduplication process on the input queries.

The dataset comprises input queries (e.g., questions about particular legal use cases) accompanied by LLM-generated responses and retrieved context data.
During development, legal professionals verified these responses to ensure they were grounded in the context provided to the LLM (as part of the prompt).
The context data is derived from a retrieval system with access to the \emph{Casetext} database for legal research, which includes case law, statutes, regulations, and legal texts authored by internal legal experts and lawyers.

The ground truth responses (LLM-based answers) were generated using custom instructions in a prompt to \emph{GPT-4} in the current production environment.
At this stage of the dataset creation process, we had compiled a selection of legal user queries, gold responses, and their corresponding contexts.

\subsection{Synthetic Adaptation}

The next step in our dataset creation process involved generating evoked ungrounded responses to evaluate both grounded and ungrounded outputs.
We instructed \emph{GPT-4o} to make subtle and unintrusive variations to the original grounded responses, preserving most of the meaning while introducing minor deviations from the provided context.
In the prompt, we included the original query and context alongside the gold response and these instructions.

These adapted responses, which we consider partially ungrounded\footnote{Only some sentences ended up with slight modifications, while most were kept as the original sentences.}, complement our final dataset.
The inclusion of both grounded and ungrounded responses allows for a more comprehensive evaluation of response quality and adherence to provided context.
An example of this subtle deviation from the source material in the generated response was depicted in the leading example in Fig. \ref{fig:example}.

\subsection{Data Splits}

We divided the dataset into training, development, and test sets using a ratio of $70:10:20$, respectively.
This split ensures a representative distribution across all subsets while maintaining a sufficiently large test set for robust evaluation.

The resulting counts for each split are presented in Table \ref{data_set_statistics}.
It is noteworthy that the number of responses is not exactly twice the number of queries.
This discrepancy arises from our dataset creation process, where we retained multiple significant variations of generated responses for certain queries to enhance the diversity and coverage of our dataset.

To maintain the integrity of our evaluation, we ensured that all responses corresponding to a particular query were assigned to the same split.
This approach prevents potential leakage between the training and evaluation sets, thereby providing a more accurate assessment of model performance on unseen data.

\section{Benchmarking Methodologies}

This section overviews the diverse methodologies employed in our benchmark study for quantifying response grounding, systematically evaluating approaches that assess adherence of generated responses to provided context.

\subsection{Similarity-based Approaches}
\label{sec:semantic_similarity}

Similarity-based approaches compare each response sentence against all context sentences, allowing for detailed grounding assessment.
We aggregate these sentence-level estimations for the final response-level prediction.

\paragraph{Semantic Similarity}
We embedded sentences using the \emph{nlpaueb/legal-bert-base-uncased} model with the Sentence-Transformers library.
Matching pairs were identified using cosine similarity, with an optimized threshold determined on the development set for final grounding prediction.

\paragraph{Quoted Information Precision}
Adapting the \emph{QuIP-score} \cite{weller-etal-2024-according}, we examined character $n$-gram overlap between LLM responses and context sentences.
We optimized both the $n$-gram size (21 in our setup) and similarity threshold on the development set for grounding determination in the final evaluation.

\subsection{Natural Language Inference}
\label{sec:nli}

\paragraph{FactKB}
Evaluating factual consistency in natural language generation is crucial, especially for complex domains.
We employed FactKB\footnote{\url{https://hf.co/bunsenfeng/FactKB}}, an approach leveraging pre-training with facts from external knowledge bases, to address challenges in entity and relation errors \cite{feng2023factkb}.

\emph{FactKB} has shown state-of-the-art performance in factual consistency evaluation across various domains.
We used it to compute factuality scores of generated response sentences against source context sentences.

Our grounding determination process involved identifying the highest-scoring source sentence for each target sentence based on \emph{FactKB} scores, then applying an optimized threshold to classify grounding sufficiency.
This threshold, determined using our development set, balanced precision and recall in grounding classification, adapting \emph{FactKB} to our specific task of response grounding quantification.

\paragraph{Hallucination Evaluation Models}
The Hallucination Evaluation Model (HEM), developed by Vectara \cite{Hughes_Vectara_Hallucination_Leaderboard_2023}, is designed to detect hallucinations in LLM-generated responses.
HEM is available in two versions: \emph{V1}, a fine-tuned model based on \emph{cross-encoder/nli-deberta-v3-base}, and \emph{V2}, an improved version using \emph{flan-t5-base}.

Built on research in factual consistency for summarization, HEM classifies whether a summary is factually consistent with its source.
The model was fine-tuned on diverse documents to ensure robustness across content types and is publicly available on Hugging Face under the Apache 2 license.

HEM evaluates LLM responses by comparing them to source documents, classifying summaries as consistent or inconsistent.
For our study, we implemented a fine-grained approach, scoring individual sentences against corresponding contexts.
This granular analysis provides a nuanced assessment of hallucinations at the sentence level, offering deeper insights into model performance.

\subsection{Prompting Approaches}

\paragraph{Direct Prompting} 
One straight-forward approach for groundedness classification via prompting is asking either the same or another LLM whether a particular response for a query is grounded in a context or not \cite{trautmann2022legal}.
Therefore, we utilized several LLMs with a custom prompt and collected the binary classification as the prompt-based baselines.
We used the specialized open access model \emph{Lynx-v1.1} \cite{ravi2024lynx} and the general purpose public LLMs \emph{GPT-4o} and \emph{Claude Sonnet 3.5}.
All three LLMs were evaluated with the same prompt from \newcite{ravi2024lynx}.

In principle, this approach has similarities with \emph{Reflexion} by \newcite{shinn2024reflexion}, where a \emph{Self-Reflection} LLM should reflect on a previous answer and if necessary to update its prediction.
The authors showed that this was helpful, especially for more complex tasks.

\paragraph{Amazon RefChecker}
RefChecker \cite{hu2024refchecker} introduces a framework for hallucination detection using knowledge triplets to capture fine-grained assertions.
The process involves three steps: claim extraction, hallucination checking, and aggregation.
This decoupled process is also known as prompt chaining \cite{trautmann2023chaining}.

An LLM identifies knowledge triplets from the response to the original query.
Zero-shot checkers then predict hallucination labels for each triplet (entailment, contradiction, or neutral).
Finally, these labels are integrated to compute an overall hallucination score for the response.

RefChecker's computational demands are notable: for n triplets extracted, the LLM is prompted with the entire original context n times, significantly impacting processing time and resource consumption.
This approach balances granular analysis with computational intensity, offering a detailed but resource-intensive method for hallucination detection.

\paragraph{SelfCheckGPT}
We adapt the approach of \newcite{manakul2023selfcheckgpt}, which assesses hallucination likelihood in LLM-generated sentences by evaluating their consistency with multiple answers from the same query.
SelfCheckGPT assumes that grounded sentences should be consistent with other sampled answers.

The method generates new responses using the initial prompt with increased temperature.
It then calculates a hallucination score for each sentence as the average of contradiction probabilities with these new samples.
The response-level score is the maximum of sentence-level scores, with the threshold optimized on the training set.

We enhance this approach with a novel context-based evaluation (\emph{ContextNLI}) using the \emph{potsawee/deberta-v3-large-mnli} model.
This compares each answer sentence against context sentences, identifying the minimum contradiction score as the hallucination probability.
The maximum score across all sentences represents the answer's overall hallucination likelihood.

We implement two variants of this approach: \emph{Multi-Gen}, which follows the original consistency checks, and our novel \emph{ContextNLI}, which incorporates the context-based evaluation, thus providing complementary methods for assessing the groundedness of LLM-generated content.

\paragraph{DeepEval: Claims Extraction and Verification}
We adapt the \textit{Faithfulness} metrics from \newcite{deepeval2023} to detect contradictions between source documents and generated answers. This approach divides the task into two subtasks: claims extraction and claim verification (prompt chains,  \newcite{trautmann2023chaining}).

First, we use an LLM to extract claims independently from both source documents and generated answers using a custom prompt. Then, a second LLM call with another custom prompt identifies claims from the generated answer not factually supported by the source document claims. If any generated claim contradicts a source claim, we consider the answer inaccurate.

This method requires three LLM calls in total: two for claims extraction and one for comparison. We utilize \emph{Claude Sonnet 3.5} for all these calls, balancing task complexity reduction with comprehensive analysis.

\subsection{Fine-Tuning}
In addition to our primary methods, we fine-tuned a Cross-Encoder classifier (\emph{DeBERTa v3} as the base model) specifically tailored to our dataset.
To ensure the integrity of our evaluation, we meticulously prepared a specialized training and evaluation corpus based on the initial data splits, thereby avoiding any potential contamination between sets.

Our fine-tuning approach focused on the nuanced differences between grounded and ungrounded responses.
For each pair of such responses, we isolated the sentences that differed between them.
This selective process allowed us to concentrate on the most informative elements for distinguishing between grounded and ungrounded content.

To establish ground truth for the grounded responses, we employed a semantic similarity measure (as described in Section \ref{sec:semantic_similarity}).
For each sentence in the grounded response, we identified the most semantically similar sentence from the context and assigned it the corresponding cosine similarity score.
These scores typically ranged from $0.8$ to $0.99$, indicating high levels of semantic alignment.

Conversely, for the ungrounded responses, we paired each sentence with the same context sentence used for its grounded counterpart.
However, we assigned these pairs a score of $1$ minus the cosine similarity, effectively inverting the grounding measure.
This approach provided a balanced representation of both grounded and ungrounded examples in our training data.

Through this methodology, we compiled a balanced dataset comprising $558$ samples for training and $75$ for development.
This carefully curated dataset served as the foundation for our fine-tuning process, enabling the Cross-Encoder to learn the subtle distinctions between grounded and ungrounded content within our specific corpus.

The outcomes of our fine-tuning efforts (after hyper-parameter optimization), are comprehensively presented (macro averaged) in Tab. \ref{tab:deberta-performance}.

\begin{table}[ht]
    \centering
    \small
    \begin{tabular}{lcccc}
    \toprule
    \textbf{Model Name} & \textbf{M-Prec} & \textbf{M-Rec} & \textbf{M-F1} & \textbf{Acc} \\
    \midrule
    deberta-v3-base & 0.459 & 0.466 & 0.450 & 0.493 \\
    deberta-v3-large & 0.736 & 0.739 & 0.733 & 0.733 \\
    \bottomrule
    \end{tabular}
    \caption{DEV set metrics for DeBERTa models}
    \label{tab:deberta-performance}
\end{table}

Following the fine-tuning stage, we integrated this grounding classification (GC) model into our benchmark, employing a methodology analogous to that used for the NLI approaches described in Section \ref{sec:nli}.

\section{Experimental Set-Up}

Our benchmarking study aimed to evaluate various methods for classifying LLM responses as grounded or ungrounded relative to a given context and query.

\paragraph{Methodology}
Despite the varied granularity of approaches (response-level vs. sentence-level), we standardized outputs to binary classifications for consistent comparison.
We developed each method on the training set, optimized parameters on the development set, and conducted final evaluations on the test set.

\paragraph{Performance Metrics}
We assessed classification accuracy (including macro-averaged f1, precision, and recall) and computational efficiency through latency measurements.
Latency was computed as the average processing time across all samples in the development set.
These metrics provide insights into each approach's practical applicability.

\paragraph{Computational Resources}
Local approaches utilized Amazon EC2 G5 Instances (8xlarge)\footnote{\url{https://aws.amazon.com/ec2/instance-types/g5/}}. Prompting-based methods were executed via Azure OpenAI Services\footnote{\url{https://azure.microsoft.com/en-us/products/ai-services/openai-service}}, AWS Bedrock (Anthropic's Claude)\footnote{\url{https://aws.amazon.com/bedrock/claude/}}, and Anthropic's API directly, ensuring diverse and robust evaluation environments.

\begin{table*}[htbp]
    \centering
    \resizebox{0.95\textwidth}{!}{
        \begin{tabular}{r|l|cccc|cccc}
        \hline
        \multirow{2}{*}{\#} & \multirow{2}{*}{Model Name} & \multicolumn{4}{c|}{Development Set} & \multicolumn{4}{c}{Test Set} \\
        \cline{3-10}
         &  & Precision & Recall & Macro-F1 & Accuracy & Precision & Recall & Macro-F1 & Accuracy \\
        \hline
         1 & COS\_SIM               & 0.525 & 0.520 & 0.494 & 0.520 & 0.497 & 0.497 & 0.493 & 0.497 \\
         2 & QUIP                   & 0.648 & 0.533 & 0.421 & 0.533 & 0.560 & 0.509 & 0.379 & 0.509 \\
        \hline
         3 & HEM V1                 & 0.640 & 0.640 & 0.640 & 0.640 & 0.598 & 0.595 & 0.592 & 0.595 \\
         4 & HEM V2                 & 0.580 & 0.580 & 0.580 & 0.580 & 0.564 & 0.563 & 0.562 & 0.563 \\
         5 & FACT\_KB               & 0.527 & 0.527 & 0.526 & 0.527 & 0.510 & 0.510 & 0.508 & 0.510 \\
         6 & GC-large               & 0.694 & 0.667 & 0.655 & 0.667 & 0.628 & 0.620 & 0.615 & 0.620 \\
        \hline
         7 & LYNX v1.1              & 0.764 & 0.460 & 0.571 & 0.460 & 0.792 & 0.503 & 0.597 & 0.503 \\
         8 & Sonnet 3.5             & 0.728 & 0.727 & 0.726 & 0.727 & 0.724 & 0.715 & 0.712 & 0.715 \\
         9 & GPT-4o                 & \underline{0.783} & \underline{0.773} & \underline{0.771} & \underline{0.773} & \textbf{0.802} & \underline{0.763} & \underline{0.755} & \underline{0.763} \\
        \hline
        10 & RefChecker (Haiku)     & 0.511 & 0.506 & 0.450 & 0.506 & 0.514 & 0.507 & 0.435 & 0.508 \\
        11 & RefChecker (Sonnet 3)  & 0.500 & 0.500 & 0.366 & 0.500 & 0.500 & 0.500 & 0.386 & 0.500 \\
        12 & DeepEval Claims Verify & \textbf{0.801} & \textbf{0.800} & \textbf{0.800} & \textbf{0.800} & \underline{0.779} & \textbf{0.774} & \textbf{0.774} & \textbf{0.775} \\
        13 & SCGPT (Multi-Gen)      & 0.627 & 0.627 & 0.627 & 0.627 & 0.679 & 0.667 & 0.661 & 0.667 \\
        14 & SCGPT (ContextNLI)     & 0.620 & 0.620 & 0.620 & 0.620 & 0.610 & 0.604 & 0.600 & 0.604 \\
        \hline
        \end{tabular}
    }
    \caption{Performance comparison of different models on Development and Test sets}
    \label{tab:model_comparison}
\end{table*}


\begin{figure*}[t!]
    \centering
    \includegraphics[width=0.7\textwidth]{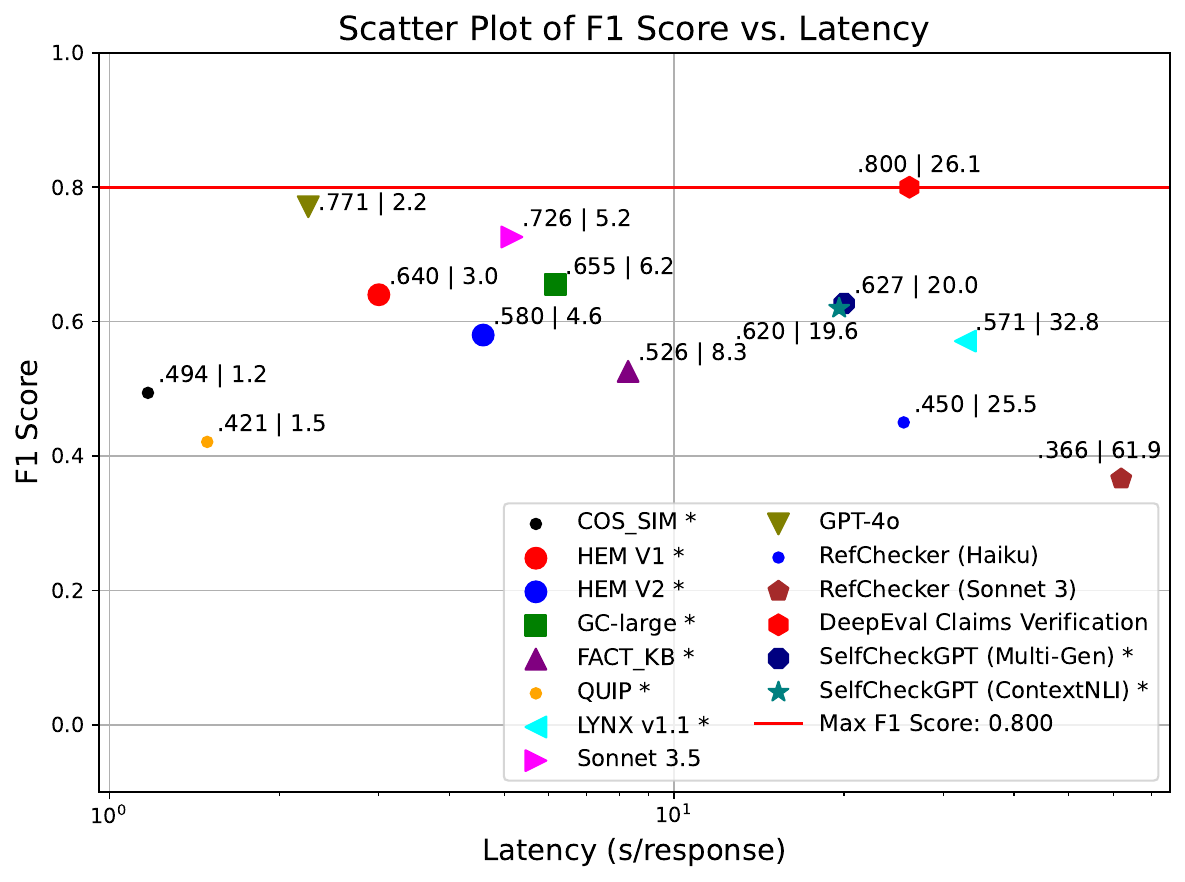}
    \caption{Development set results for our benchmark. We report the F1-scores (y-axis) for each method and the corresponding latency (x-axis) in seconds per response. Approach names denoted with \emph{*} were run on an AWS \emph{ml.8xlarge} instance.}
    \label{fig:results_plot}
\end{figure*}


\begin{figure*}[htbp]
  \centering
  \includegraphics[width=1.0\textwidth]{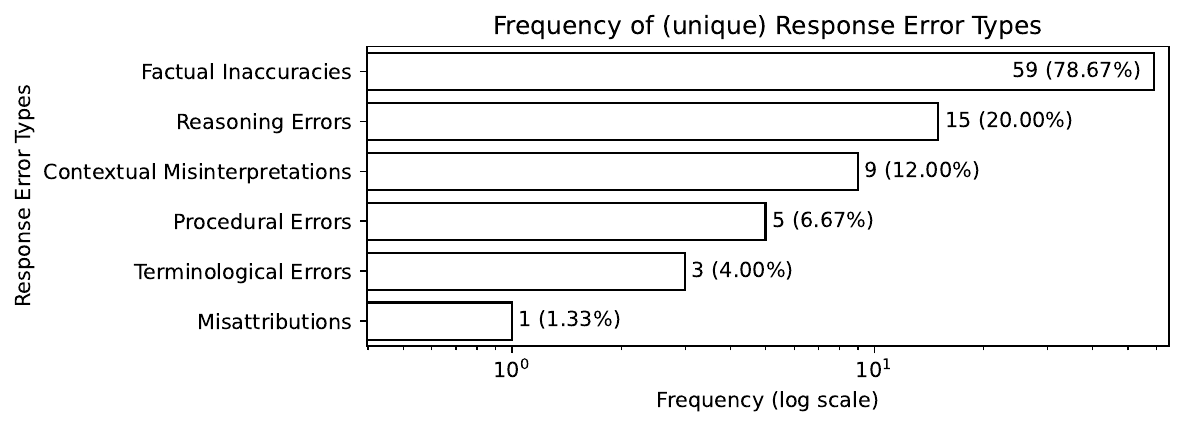}
  \caption{Counts of unique error types in the development set. Some responses contained up to three different error types. The frequency axis is in log-scale.}
  \label{fig:unique_error_types}
\end{figure*}


\begin{table*}[t]
    \centering
    \resizebox{0.65\textwidth}{!}{
        \begin{tabular}{l|r|r|r}
            \hline
            \textbf{Error Type} & \textbf{Misclassified} & \textbf{Total} & \textbf{Percentage} \\
            \hline
            Terminological Errors         &  2 &  3 & 66.7\% \\
            Factual Inaccuracies          & 12 & 59 & 20.3\% \\
            Procedural Errors             &  1 &  5 & 20.0\% \\
            Reasoning Errors              &  2 & 15 & 13.3\% \\
            Contextual Misinterpretations &  1 &  9 & 11.1\% \\
            Misattributions               &  0 &  1 &  0.0\% \\
            \hline
        \end{tabular}
    }
    \caption{Development set misclassification of the best performing model by error types.}
    \label{tab:misclassificaiton}
\end{table*}

\section{Groundedness Classification Results}

Our benchmark evaluation of groundedness classification approaches revealed insightful performance trade-offs, as shown in Tab. \ref{tab:model_comparison}.
The metrics include classification precision, recall, F1-score, and accuracy, providing a comprehensive view of each method's applicability.

The multi-stage prompt chaining approach, \emph{DeepEval Claims Verify}, achieved top classification metrics, but with high latency ($26.1$ seconds per request).
In contrast, \emph{direct prompting} with \emph{GPT-4o} achieved the second-highest scores with significantly lower latency ($2.2$ seconds), as illustrated in Fig. \ref{fig:results_plot}.

A clear speed-performance trade-off emerged across methods.
Similarity-based approaches (\emph{COS\_SIM} and \emph{QUIP}) were fastest but struggled with ungrounded response identification.
NLI methods showed improved performance at the cost of increased latency.
Within NLI, \emph{HEM V1} outperformed \emph{HEM V2}, and fine-tuning on our corpus further improving results.

Unexpectedly, complex prompt chaining approaches like \emph{RefChecker} and \emph{SelfCheckGPT} underperformed, highlighting challenges in developing universally effective methods across diverse contexts.

These findings emphasize the importance of balancing task performance and computational efficiency when selecting a groundedness classification approach, with optimal choices depending on specific application requirements and resource constraints.

\section{Error Analysis}

We conducted a detailed investigation into the types of response errors present in our benchmark dataset to gain deeper insights into ungrounded content.

Through examination of error spans in the training set, we identified six distinct error types.
The models were instructed to select from our predefined error types (Tab. \ref{tab:error_types}, App. \ref{app:error_description}).

Focusing on the development set, our analysis revealed interesting patterns.
The LLMs achieved exact agreement on the hallucination type in $29\%$ of cases, with at least one overlapping error type for each response.
\emph{GPT-4o} typically predicted a single error type, while \emph{Claude-3.5-Sonnet} often suggested multiple types per response.

We aggregated predictions where both LLMs agreed.
The distribution of unique error types is visualized in Fig. \ref{fig:unique_error_types}, with per-response occurrences in Fig. \ref{fig:response_error_types} (App. \ref{app:dev_set_errors}).
\emph{Factual Inaccuracies} were most common, followed by \emph{Reasoning Errors}.
All initially defined error types were represented, validating our classification scheme.

This analysis provides valuable insights into response error types and ungrounded content in language model outputs, crucial for developing targeted strategies to improve response generation.

\paragraph{Misclassification Analysis}
We conducted a misclassification analysis on our best-performing model, \emph{DeepEval Claims Verify}, to gain deeper insights into its performance across different error types.
As summarized in Tab. \ref{tab:misclassificaiton}, \emph{Terminological Errors} showed the highest misclassification rate ($67\%$), despite their low frequency, followed by \emph{Factual Inaccuracies} ($20\%$) and \emph{Procedural Errors} ($20\%$).
These findings reveal the varying challenges posed by different error categories and highlight areas for potential improvement in groundedness classification models, particularly in handling less common but difficult-to-classify error types.

\section{Conclusion}
\label{sec:conclusion}

Our comprehensive benchmark study on groundedness classification of legal question-answering systems has revealed significant insights into performance and efficiency trade-offs.
The multi-stage prompt chaining approach, \emph{DeepEval Claims Verify}, emerged as the top performer with an F1 score of $0.80$, closely followed by direct prompting using \emph{GPT-4o} at $0.77$, which demonstrated lower latency.
These results highlight the potential of advanced prompting techniques in achieving high accuracy.

Similarity-based and natural language inference methods, while less accurate, offered fast processing times.
Our response error type classification identified \emph{Factual Inaccuracies} and \emph{Reasoning Errors} as the most prevalent types of ungrounded content, providing direction for future improvements.

The study underscores the critical balance between task performance, computational efficiency, and ease of implementation when selecting groundedness classification methods.
With top-performing methods achieving F1 scores of $0.80$, this benchmark represents a significant advancement in the reliable assessment of AI-generated content across diverse applications.

\newpage
\section*{Limitations}

While our study offers valuable insights into the performance of various groundedness classification approaches, it is essential to acknowledge several limitations inherent in our experimental setup and the methods we evaluated.

Firstly, our dataset, though carefully curated, is limited in size and domain scope.
The responses were generated using specific language models and may not fully represent the diverse range of hallucinations or ungrounded content that could occur across different models or domains.
This limitation potentially affects the generalizability of our findings to broader contexts or more specialized applications.

Secondly, the binary classification of responses as either grounded or ungrounded may oversimplify the nuanced nature of language model outputs.
In reality, responses often contain a mix of grounded and ungrounded elements, and a more granular assessment might provide deeper insights into model behavior.

Our evaluation metrics, while standard in the field, may not capture all aspects of response quality or usefulness.
For instance, a response that is technically grounded but irrelevant or poorly structured might still receive a high rating within our current framework.

The computational resources required for some of the more complex approaches, particularly those involving multiple API calls or large language models, pose scalability challenges.
This limitation may restrict the practical applicability of these methods in real-time or resource-constrained environments.

Additionally, our error type classification, while informative, relies on the agreement between two specific language models.
This approach may introduce biases or limitations based on the particular characteristics of these models.

Lastly, the rapid pace of development in language model technology means that our findings may quickly become outdated as new models and techniques emerge.
The performance gaps we observed between different approaches may shift with the introduction of more advanced models or refined methodologies.

Future work should address these limitations by expanding the dataset to include a broader range of domains and increasing its size.
Developing more nuanced classification frameworks that can capture the complexity of language model outputs would also be beneficial.
Furthermore, exploring scalable methods that can be applied in real-time or resource-constrained environments, as well as continuously updating the evaluation framework to reflect the latest advancements in language model technology, will be crucial for the ongoing relevance of this research.

\section*{Ethics Statement}

This study on groundedness classification methods aims to improve the reliability and trustworthiness of AI-generated content, which has significant ethical implications.
By developing more accurate methods to detect ungrounded or hallucinated information, we contribute to the broader goal of mitigating the spread of misinformation and enhancing the integrity of AI-assisted communication.
Our work aligns with the principles of beneficence and non-maleficence, as it seeks to maximize the benefits of language models while minimizing potential harms associated with inaccurate or misleading information.

We acknowledge that the development and deployment of these classification methods may have broader societal impacts.
We emphasize the importance of transparent and responsible use of these methods, respecting principles of fairness and privacy.
Furthermore, we encourage ongoing dialogue and collaboration within the NLP community to address the ethical challenges associated with AI-generated content and its evaluation.

\bibliography{custom}

\begin{thebibliography}{25}
\expandafter\ifx\csname natexlab\endcsname\relax\def\natexlab#1{#1}\fi

\bibitem[{Bubeck et~al.(2023)Bubeck, Chandrasekaran, Eldan, Gehrke, Horvitz, Kamar, Lee, Lee, Li, Lundberg et~al.}]{bubeck2023sparks}
S{\'e}bastien Bubeck, Varun Chandrasekaran, Ronen Eldan, Johannes Gehrke, Eric Horvitz, Ece Kamar, Peter Lee, Yin~Tat Lee, Yuanzhi Li, Scott Lundberg, et~al. 2023.
\newblock Sparks of artificial general intelligence: Early experiments with gpt-4. arxiv.
\newblock \emph{arXiv preprint arXiv:2303.12712}.

\bibitem[{Chandu et~al.(2021)Chandu, Bisk, and Black}]{chandu-etal-2021-grounding}
Khyathi~Raghavi Chandu, Yonatan Bisk, and Alan~W Black. 2021.
\newblock \href {https://doi.org/10.18653/v1/2021.findings-acl.375} {Grounding {`}grounding{'} in {NLP}}.
\newblock In \emph{Findings of the Association for Computational Linguistics: ACL-IJCNLP 2021}, pages 4283--4305, Online. Association for Computational Linguistics.

\bibitem[{Feng et~al.(2023)Feng, Balachandran, Bai, and Tsvetkov}]{feng2023factkb}
Shangbin Feng, Vidhisha Balachandran, Yuyang Bai, and Yulia Tsvetkov. 2023.
\newblock Factkb: Generalizable factuality evaluation using language models enhanced with factual knowledge.
\newblock In \emph{Proceedings of the 2023 Conference on Empirical Methods in Natural Language Processing}, pages 933--952.

\bibitem[{Hong et~al.(2024)Hong, Gema, Saxena, Du, Nie, Zhao, Perez-Beltrachini, Ryabinin, He, and Minervini}]{hong2024hallucinations}
Giwon Hong, Aryo~Pradipta Gema, Rohit Saxena, Xiaotang Du, Ping Nie, Yu~Zhao, Laura Perez-Beltrachini, Max Ryabinin, Xuanli He, and Pasquale Minervini. 2024.
\newblock The hallucinations leaderboard--an open effort to measure hallucinations in large language models.
\newblock \emph{arXiv preprint arXiv:2404.05904}.

\bibitem[{Hu et~al.(2024)Hu, Ru, Qiu, Guo, Zhang, Xu, Luo, Liu, Zhang, and Zhang}]{hu2024refchecker}
Xiangkun Hu, Dongyu Ru, Lin Qiu, Qipeng Guo, Tianhang Zhang, Yang Xu, Yun Luo, Pengfei Liu, Yue Zhang, and Zheng Zhang. 2024.
\newblock Refchecker: Reference-based fine-grained hallucination checker and benchmark for large language models.
\newblock \emph{arXiv preprint arXiv:2405.14486}.

\bibitem[{Hughes et~al.(2023)Hughes, Bae, and Li}]{Hughes_Vectara_Hallucination_Leaderboard_2023}
Simon Hughes, Minseok Bae, and Miaoran Li. 2023.
\newblock \href {https://github.com/vectara/hallucination-leaderboard} {{Vectara Hallucination Leaderboard}}.

\bibitem[{Ip(2023)}]{deepeval2023}
Jeffrey Ip. 2023.
\newblock \href {https://github.com/confident-ai/deepeval} {Deepeval: A tool for deep learning model evaluation}.
\newblock GitHub repository.

\bibitem[{Khazaeli et~al.(2021)Khazaeli, Punuru, Morris, Sharma, Staub, Cole, Chiu-Webster, and Sakalley}]{khazaeli2021free}
Soha Khazaeli, Janardhana Punuru, Chad Morris, Sanjay Sharma, Bert Staub, Michael Cole, Sunny Chiu-Webster, and Dhruv Sakalley. 2021.
\newblock A free format legal question answering system.
\newblock In \emph{Proceedings of the Natural Legal Language Processing Workshop 2021}, pages 107--113.

\bibitem[{Kry{\'s}ci{\'n}ski et~al.(2020)Kry{\'s}ci{\'n}ski, McCann, Xiong, and Socher}]{kryscinski2020evaluating}
Wojciech Kry{\'s}ci{\'n}ski, Bryan McCann, Caiming Xiong, and Richard Socher. 2020.
\newblock Evaluating the factual consistency of abstractive text summarization.
\newblock In \emph{Proceedings of the 2020 Conference on Empirical Methods in Natural Language Processing (EMNLP)}, pages 9332--9346.

\bibitem[{Li et~al.(2023)Li, Zhao, Chia, Ding, Joty, Poria, and Bing}]{li2023chain}
Xingxuan Li, Ruochen Zhao, Yew~Ken Chia, Bosheng Ding, Shafiq Joty, Soujanya Poria, and Lidong Bing. 2023.
\newblock Chain-of-knowledge: Grounding large language models via dynamic knowledge adapting over heterogeneous sources.
\newblock In \emph{The Twelfth International Conference on Learning Representations}.

\bibitem[{Magesh et~al.(2024)Magesh, Surani, Dahl, Suzgun, Manning, and Ho}]{magesh2024hallucination}
Varun Magesh, Faiz Surani, Matthew Dahl, Mirac Suzgun, Christopher~D Manning, and Daniel~E Ho. 2024.
\newblock Hallucination-free? assessing the reliability of leading ai legal research tools.
\newblock \emph{arXiv preprint arXiv:2405.20362}.

\bibitem[{Manakul et~al.(2023)Manakul, Liusie, and Gales}]{manakul2023selfcheckgpt}
Potsawee Manakul, Adian Liusie, and Mark Gales. 2023.
\newblock Selfcheckgpt: Zero-resource black-box hallucination detection for generative large language models.
\newblock In \emph{Proceedings of the 2023 Conference on Empirical Methods in Natural Language Processing}, pages 9004--9017.

\bibitem[{Martinez-Gil(2023)}]{martinez2023survey}
Jorge Martinez-Gil. 2023.
\newblock A survey on legal question--answering systems.
\newblock \emph{Computer Science Review}, 48:100552.

\bibitem[{Maynez et~al.(2020)Maynez, Narayan, Bohnet, and McDonald}]{maynez-etal-2020-faithfulness}
Joshua Maynez, Shashi Narayan, Bernd Bohnet, and Ryan McDonald. 2020.
\newblock \href {https://doi.org/10.18653/v1/2020.acl-main.173} {On faithfulness and factuality in abstractive summarization}.
\newblock In \emph{Proceedings of the 58th Annual Meeting of the Association for Computational Linguistics}, pages 1906--1919, Online. Association for Computational Linguistics.

\bibitem[{Monroy et~al.(2009)Monroy, Calvo, and Gelbukh}]{monroy2009nlp}
Alfredo Monroy, Hiram Calvo, and Alexander Gelbukh. 2009.
\newblock Nlp for shallow question answering of legal documents using graphs.
\newblock In \emph{International Conference on Intelligent Text Processing and Computational Linguistics}, pages 498--508. Springer.

\bibitem[{Ravi et~al.(2024)Ravi, Mielczarek, Kannappan, Kiela, and Qian}]{ravi2024lynx}
Selvan~Sunitha Ravi, Bartosz Mielczarek, Anand Kannappan, Douwe Kiela, and Rebecca Qian. 2024.
\newblock Lynx: An open source hallucination evaluation model.
\newblock \emph{arXiv preprint arXiv:2407.08488}.

\bibitem[{Rawte et~al.(2023)Rawte, Sheth, and Das}]{rawte2023survey}
Vipula Rawte, Amit Sheth, and Amitava Das. 2023.
\newblock A survey of hallucination in large foundation models.
\newblock \emph{arXiv preprint arXiv:2309.05922}.

\bibitem[{Rosenthal et~al.(2024)Rosenthal, Sil, Florian, and Roukos}]{rosenthal2024clapnq}
Sara Rosenthal, Avirup Sil, Radu Florian, and Salim Roukos. 2024.
\newblock Clapnq: Cohesive long-form answers from passages in natural questions for rag systems.
\newblock \emph{arXiv preprint arXiv:2404.02103}.

\bibitem[{Shinn et~al.(2024)Shinn, Cassano, Gopinath, Narasimhan, and Yao}]{shinn2024reflexion}
Noah Shinn, Federico Cassano, Ashwin Gopinath, Karthik Narasimhan, and Shunyu Yao. 2024.
\newblock Reflexion: Language agents with verbal reinforcement learning.
\newblock \emph{Advances in Neural Information Processing Systems}, 36.

\bibitem[{Trautmann(2023)}]{trautmann2023chaining}
Dietrich Trautmann. 2023.
\newblock \href {https://arxiv.org/abs/2308.04138} {Large language model prompt chaining for long legal document classification}.
\newblock \emph{arXiv preprint arXiv:2308.04138}.

\bibitem[{Trautmann et~al.(2022)Trautmann, Petrova, and Schilder}]{trautmann2022legal}
Dietrich Trautmann, Alina Petrova, and Frank Schilder. 2022.
\newblock \href {https://arxiv.org/abs/2212.02199} {Legal prompt engineering for multilingual legal judgement prediction}.
\newblock \emph{arXiv preprint arXiv:2212.02199}.

\bibitem[{Vold and Conrad(2021)}]{vold2021using}
Andrew Vold and Jack~G Conrad. 2021.
\newblock Using transformers to improve answer retrieval for legal questions.
\newblock In \emph{Proceedings of the Eighteenth International Conference on Artificial Intelligence and Law}, pages 245--249.

\bibitem[{Wang et~al.(2024)Wang, Duan, Li, Wang, and Cai}]{wang2024llms}
Keheng Wang, Feiyu Duan, Peiguang Li, Sirui Wang, and Xunliang Cai. 2024.
\newblock Llms know what they need: Leveraging a missing information guided framework to empower retrieval-augmented generation.
\newblock \emph{arXiv preprint arXiv:2404.14043}.

\bibitem[{Weller et~al.(2024)Weller, Marone, Weir, Lawrie, Khashabi, and Van~Durme}]{weller-etal-2024-according}
Orion Weller, Marc Marone, Nathaniel Weir, Dawn Lawrie, Daniel Khashabi, and Benjamin Van~Durme. 2024.
\newblock \href {https://aclanthology.org/2024.eacl-long.140} {{``}according to . . . {''}: Prompting language models improves quoting from pre-training data}.
\newblock In \emph{Proceedings of the 18th Conference of the European Chapter of the Association for Computational Linguistics (Volume 1: Long Papers)}, pages 2288--2301, St. Julian{'}s, Malta. Association for Computational Linguistics.

\bibitem[{Zhu et~al.(2024)Zhu, Sun, and Yang}]{zhu2024halueval}
Zhiying Zhu, Zhiqing Sun, and Yiming Yang. 2024.
\newblock Halueval-wild: Evaluating hallucinations of language models in the wild.
\newblock \emph{arXiv preprint arXiv:2403.04307}.

\end{thebibliography}
\bibliographystyle{acl_natbib}

\appendix

\section{Response Error Types}

\subsection{Dev Set Error Types}
\label{app:dev_set_errors}

\begin{figure*}[htbp]
  \centering
  \includegraphics[width=1.0\textwidth]{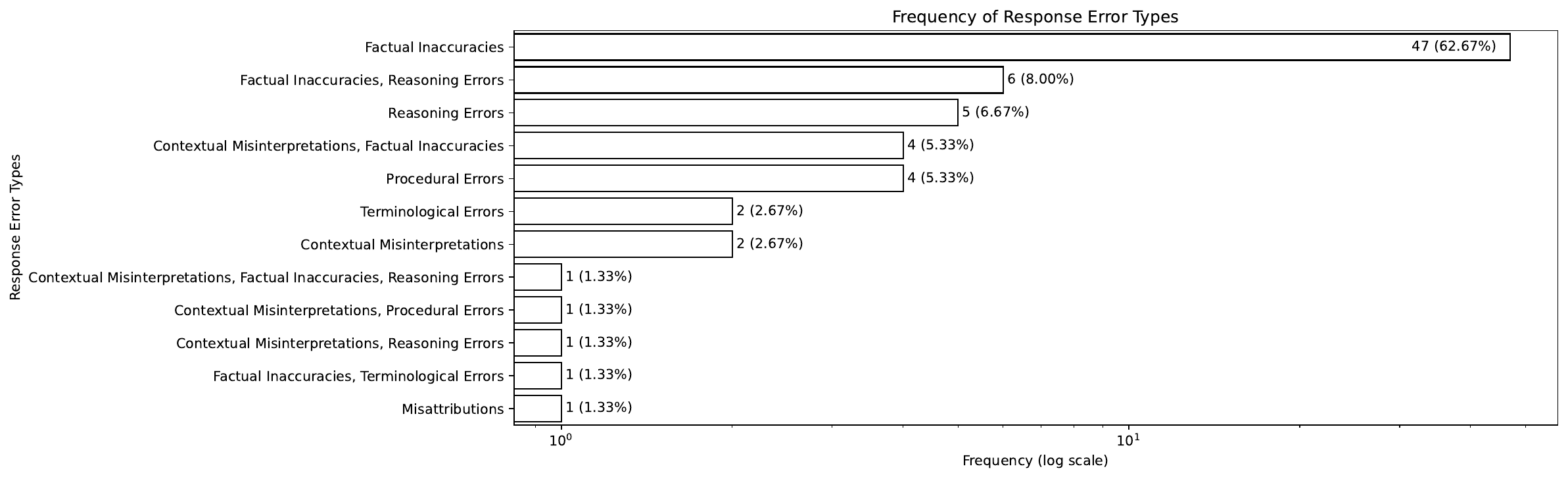}
  \caption{Counts of response error types in the development set. The frequency axis is in log-scale.}
  \label{fig:response_error_types}
\end{figure*}

\subsection{Description and Examples}
\label{app:error_description}
See the table \ref{tab:error_types} for our six response error types with their descriptions and examples.

\begin{table*}[h]
\centering
    \begin{tabular}{|p{0.2\textwidth}|p{0.3\textwidth}|p{0.4\textwidth}|}
    \hline
    \textbf{Error Type} & \textbf{Short Description} & \textbf{Examples} \\
    \hline
    Factual Inaccuracies & Misrepresentation of established facts, dates, or details & 1. Brown v. Board of Education was decided in 1964. \newline 2. The First Amendment protects only written speech. \\
    \hline
    Contextual Misinterpretations & Misapplication of legal principles or inappropriate analogies & 1. Applying Miranda rights to a civil tax dispute. \newline 2. Using Roe v. Wade precedent in a Second Amendment case. \\
    \hline
    Procedural Errors & Mistakes in describing legal procedures or processes & 1. A case goes directly from district court to the Supreme Court, skipping the appellate court. \newline 2. Claiming that jury selection occurs after opening statements in a trial. \\
    \hline
    Reasoning Errors & Flawed arguments or unsupported legal conclusions & 1. Since the Fourth Amendment protects against unreasonable searches, all warrantless searches are unconstitutional. \newline 2. Because the Supreme Court ruled on abortion in Roe v. Wade, states cannot pass any abortion laws. \\
    \hline
    Misattributions & Incorrect assignment of opinions, quotes, or actions & 1. Justice Scalia wrote the majority opinion in Obergefell v. Hodges. \newline 2. The phrase "separate but equal" originated from Brown v. Board of Education. \\
    \hline
    Terminological Errors & Misuse or misinterpretation of legal terms or concepts & 1. "Habeas corpus" refers to the right to a speedy trial. \newline 2. "Strict scrutiny" means that a law is automatically unconstitutional. \\
    \hline
    \end{tabular}
\caption{Response error types with a description and examples}
\label{tab:error_types}
\end{table*}

\end{document}